\documentclass[conference]{IEEEtran}
\IEEEoverridecommandlockouts
\usepackage{cite}
\usepackage[numbers]{natbib}

\usepackage{amsmath,amssymb,amsfonts}
\usepackage{algorithmic}
\usepackage{graphicx}
\usepackage{subcaption}
\usepackage{textcomp}
\usepackage{float}
\usepackage{xcolor}
\usepackage[nolist]{acronym}
\usepackage{comment}
\usepackage{cite}
\usepackage{amsmath,amssymb,amsfonts}
\usepackage{algorithmic}
\usepackage{graphicx}
\usepackage{subcaption}
\usepackage{textcomp}
\usepackage{xcolor}
\usepackage[bookmarks=false]{hyperref}
\hypersetup{nolinks=true}
\usepackage{booktabs}
\usepackage{tabularx}
\usepackage{listings}
\usepackage{color}
\usepackage[autostyle=true]{csquotes}
\usepackage{tikz}
\usepackage{lipsum}
\usetikzlibrary{shapes,snakes}
\usetikzlibrary{calc} 
\usetikzlibrary{arrows.meta}
\usetikzlibrary{positioning, fit}
\tikzset{
	>=Latex,
	line/.style={draw,->},
	anode/.style={rectangle,draw,
		align=center,rounded corners,minimum height=4em,font=\strut},
	bnode/.style={anode,fill=white, font=\strut},
	cnode/.style={anode, fill=cyan!20, font=\strut},
treenode/.style = {circle,
	draw=black,thick, fill=white, align=center, minimum size=1cm},
root/.style     = {treenode, font=\footnotesize},
env/.style      = {treenode, font=\footnotesize}, 
dummy/.style    = {circle,draw}
}
\tikzstyle{decision} = [diamond, draw, fill=yellow!20, 
    text width=6em, text badly centered, node distance=3cm, inner sep=0pt, line width=0.8pt, minimum height=9em,
    minimum width=9em]
\tikzstyle{block} = [rectangle, draw, fill=cyan!20, 
    text width=6.7em, text centered, rounded corners, minimum height=4em, line width=0.8pt]
\tikzstyle{line} = [draw, -latex, line width=0.8pt]
\tikzstyle{dashline} = [draw, -latex,dashed, line width=0.8pt]
\tikzstyle{helpline} = [draw, line width=0.8pt]
\tikzstyle{method} =  [trapezium, trapezium left angle=80, trapezium right angle=-80,text centered,text width = 2cm,minimum height=1cm, minimum width=2cm, draw=black, fill=yellow!20, line width=0.8pt]
\tikzstyle{exArtifact} =[rectangle, draw, fill=cyan!20, 
    text width=6.7em, text centered, rounded corners, minimum height=4em, line width=0.8pt]
\tikzstyle{genArtifact} =[rectangle, draw, fill=green!20, 
    text width=6.7em, text centered, rounded corners, minimum height=4em, line width=0.8pt]
\newcommand{\method}[1]{\textit{#1}}

\usepackage[nolist]{acronym}
\usepackage{comment}
\def\BibTeX{{\rm B\kern-.05em{\sc i\kern-.025em b}\kern-.08em
    T\kern-.1667em\lower.7ex\hbox{E}\kern-.125emX}}

\def\BibTeX{{\rm B\kern-.05em{\sc i\kern-.025em b}\kern-.08em
    T\kern-.1667em\lower.7ex\hbox{E}\kern-.125emX}}

\begin{document}
\bstctlcite{IEEEexample:BSTcontrol}
\title{Model-based Workflow for the Automated Generation of PDDL Descriptions}

\author{
\IEEEauthorblockN{
Hamied Nabizada,
Tom Jeleniewski,
Felix Gehlhoff}

\IEEEauthorblockA{ Institute of Automation Technology\\
Helmut Schmidt University, Hamburg, Germany\\
\{hamied.nabizada, tom.jeleniewski, felix.gehlhoff\}@hsu-hh.de\\}
\and 
\IEEEauthorblockN{Alexander Fay}
\IEEEauthorblockA{Chair of Automation\\
Ruhr University,
Bochum, Germany \\
alexander.fay@rub.de}
}

\maketitle
\begin{abstract}
Manually creating Planning Domain Definition Language (PDDL) descriptions is difficult, error-prone, and requires extensive expert knowledge. 
However, this knowledge is already embedded in engineering models and can be reused. 
Therefore, this contribution presents a comprehensive workflow for the automated generation of PDDL descriptions from integrated system and product models.

The proposed workflow leverages Model-Based Systems Engineering (MBSE) to organize and manage system and product information, translating it automatically into PDDL syntax for planning purposes. 
By connecting system and product models with planning aspects, it ensures that changes in these models are quickly reflected in updated PDDL descriptions, facilitating efficient and adaptable planning processes. 
The workflow is validated within a use case from aircraft assembly.

\end{abstract}


\begin{acronym}[ECU]

\acro{bnf}[BNF]{Backus-Naur-Form}
\acro{pddl}[PDDL]{Planning Domain Definition Language}
\acro{dsml}[DSML]{Domain-Specific Modelling Language}
\acroplural{dsml}[DSMLs]{Domain-Specific Modelling Languages}
\acro{msosa}[MSoSA]{Magic Systems of Systems Architect}
\acro{sysml}[SysML]{Systems Modeling Language}
\acro{uml}[UML]{Unified Modeling Language}
\acro{vtl}[VTL]{Velocity Template Language}
\acro{xml}[XML]{Extensible Markup Language}
\acro{omg}[OMG]{Object Management Group}
\acro{hddl}[HDDL]{Hierarchical Domain Definition Language}

\acro{mbse}[MBSE]{Model-Based Systems Engineering}
\acro{uml2}[UML 2]{Unified Modeling Language 2}
\acro{uml}[UML]{Unified Modeling Language}
\acro{owl}[OWL]{Web Ontology Language}
\acro{cps}[CPS]{cyber-physical system}
\acroplural{cps}[CPSs]{cyber-physical systems}
\acro{kpi}[KPI]{key performance indicator}
\acroplural{kpi}[KPIs]{key performance indicators}
\acro{csmop}[CSMOP]{constraint satisfaction and multi-criteria objective problem}
\acro{omg}[OMG]{Object Management Group}
\acro{fwb}[FWB]{Functional Whitebox}
\acro{rflt}[RFLT]{Requirement-Functional-Logical-Technical}
\acro{spes}[SPES]{Software Platform Embedded Systems}
\acro{pddl}[PDDL]{Planning Domain Definition Language}
\acro{hse}[HSE]{Health-Safety-Environment}
\acro{ocl}[OCL]{Object Constraint Language}
\acro{xmi}[XMI]{XML Metadata Interchange}
\acro{omg}[OMG]{Object Management Group}
\acro{ppr}[PPR]{product, process and resource}
\acro{fpd}[FPD]{Formalized Process Description}
\acro{mfee}[MFEE]{multifunctional end effector}

\end{acronym}
\section{Introduction}
The rapid advancement of technology in recent years has led to increasingly complex production systems and products. 
\ac{mbse} has emerged as a powerful approach for addressing these challenges by providing a structured methodology for modeling, analyzing, and managing complex systems~\cite{Henderson.2021}. 
\ac{mbse} promotes the use of models to provide detailed and consistent descriptions of production systems, facilitating efficient collaboration among the various disciplines involved in the development process~\cite{Schmidt.2020}.

Methods of \ac{mbse}, such as the \ac[4]{sysml}-based modeling approach described by \cite{weilkiens2016sysmod}, offer structured methodologies for creating system models. 
These structured approaches reduce the heterogeneity of the contents and representations of individual system submodels, thereby promoting the reusability and comparability of the created models~\cite{Estefan.2007}. 
However, these methods often require numerous manual steps, such as assigning numerous individual process steps to potential technical resources that can execute these processes, which demands a high level of expert knowledge. 
This is particularly true when comparing different system variants to determine the ideal configuration for specific optimization goals. 

In system development, it is crucial to determine which system variant is best suited to efficiently perform a given set of processes under specific conditions~\cite{Törmanen.2017}. 
To facilitate this determination, process plans are required that define the sequence of actions to be executed and assign these actions to the corresponding subsystems. 
Automating this process planning is desirable to reduce the time required to evaluate different system variants. 
This automation requires the integration of specific planning aspects into system modeling. 
Planning aspects refer to the definition of actions necessary to achieve specific goals within the modeled system. 
These aspects are central to the field of AI planning, which focuses on the development of algorithms and languages for solving planning problems~\cite{Ghallab.2016}.

The \ac{pddl} has emerged as the de facto standard for describing planning problems~\cite{MayrDorn.2022}. 
\ac{pddl} consists of two main components: i)~the domain, which defines the types, predicates, and actions available, and ii) the problem, which specifies the initial state, goals, and constraints of the planning task. 
However, the modeling of \ac{pddl} domain descriptions is often considered particularly challenging, time-consuming, and error-prone~\cite{Lindsay.2023}. 
Similarly, the modeling of \ac{pddl} problem instances presents challenges, such as data inconsistency issues~\cite{Bhatnagar.2022}. 
This paper presents a comprehensive workflow for the automated generation of \ac{pddl} descriptions from system and product models, leveraging \ac{mbse} to enhance the planning process. 
The workflow consists of four phases: (I) \textit{Analysis and Preparation of the System Model}, (II)~\textit{Enrichment of the System Model}, (III) \textit{Provision of the Product Model}, and (IV) \textit{Generation of PDDL Descriptions}. 
By connecting the system and product models with the planning aspects, the workflow ensures that changes in these models are reflected in updated \ac{pddl} descriptions. 

The remainder of this paper is structured as follows. 
Section~\ref{sec:relatedwork} provides an overview of related work. 
Section~\ref{sec:workflow} presents the proposed workflow. 
Section~\ref{sec:applicationexample} applies the workflow to an application case in aircraft assembly. 
Finally, Section~\ref{sec:conclusionandoutlook} concludes the paper and outlines future research.

\section{Related Work}
\label{sec:relatedwork}
Huckaby et al. \cite{Huckaby.2013} introduce a \ac{sysml} taxonomy for assembly tasks to describe system capabilities, which form the basis for deriving \ac{pddl} actions. 
However, the \ac{pddl} descriptions are created manually and are limited to the taxonomy. 

Vieira da Silva et al. \cite{vieira2023transformation} present an ontology-based approach to automatically generate \ac{pddl} descriptions by matching required and offered capabilities. 
This approach requires a specific capability model and does not use existing models. 

Rimani et al. \cite{Rimani.2021} simplifies planning problems in \ac{hddl}, a \ac{pddl} dialect, by comparing HDDL and SysML elements and proposing a conceptual workflow for modeling \ac{hddl}. 
However, the transformation must be done manually.

Wally et al. \cite{wally2019flexible} describe an approach that converts ISA-95 manufacturing system models into \ac{pddl} descriptions for production planning. 
This approach is limited to ISA-95 systems and does not integrate with other models. 

Konidaris et al. \cite{Konidaris.2018} present a method for developing abstract symbolic representations to aid high-level planning in robotics, focussing on how robots can utilize these representations derived from sensorimotor data. 
This approach prioritizes learning from direct interactions over using existing models. 

Stoev et al. \cite{Stoev.2023} present a tool for automating the generation of domain-specific symbolic models from texts. 
This tool streamlines the process by extracting domain knowledge from instructional texts, like cooking recipes, and automatically generating \ac{pddl} descriptions. 
Similar to the approach described in \cite{Konidaris.2018}, this tool also does not use existing models. 

Nabizada et al. \cite{nabizada2024eka} introduce a dedicated \ac{sysml} profile that embeds \ac{pddl} constructs directly into \ac{sysml} models. 
This approach facilitates automated planning by creating an interface between system modeling and planning algorithms. 

While these approaches contribute valuable methods for generating \ac{pddl} descriptions, they often require manual intervention, lack integration with existing models, or are limited to specific frameworks. 
Therefore, there is a need for a comprehensive workflow that leverages existing system and product models, integrating them seamlessly to generate automatically \ac{pddl} descriptions. 
The following section introduces such a workflow designed to address these gaps.

\section{Workflow for Automated Generation of PDDL Descriptions}
\label{sec:workflow}
\begin{figure*}[ht]
    \centering
    \resizebox{0.8\textwidth}{!}{
\centering
\begin{tikzpicture}[node distance= 3cm]

\node[exArtifact](sysModel){System Model};
   \node [method, right of= sysModel] (analyzeSM) {Analyze System Model};
   \node [method, right of= analyzeSM] (scope) {Define Scope of Observation};
   \node [method, right of= scope] (identRelElements) {Identify relevant Elements};
   \node[exArtifact, below of= sysModel](profile){PDDL Profile};
   \node[method, right of= profile](genDomain){Create PDDL Domain (excl. Actions)};
   \node[method, right of= genDomain](defAction){Define Actions of Domain};
   \node[genArtifact, right of= defAction](aufbSM){Extended System Model};
   \node[exArtifact, below of= profile](product){Product Model};
   \node[method, right of= aufbSM](genPD){Generate PDDL Descriptions};
   \node[method, right of = product](identProdInfo){Identify Product Information};
   \node[method, right of = identProdInfo](extract){Extract Relevant Information};
   \node[method, right of = extract](transmit){Transfer to MBSE Environment};
   \node[method, right of = transmit](annot){Annotate according to PDDL Domain};
   \node[genArtifact, right of = annot](aufbPM){Extended Product Model};
   \node[exArtifact, above of= genPD](algorithm) {Algorithm};
   \node[genArtifact, right of = genPD](problem){PDDL Domain \& Problem};
    \node[method, right of = problem](solve){Solve Problem Description};
    \node[genArtifact, above of = solve](plan){PDDL Plan};
\node[draw, dotted, line width=0.8pt,fit=(sysModel)(analyzeSM)(scope)(identRelElements), inner sep=0.1cm] (phase1) {};
\node[rotate=90, above=0cm] at (phase1.west) {\textbf{Phase I}};
\node[draw, dotted, line width=0.8pt,fit=(profile)(genDomain)(defAction)(aufbSM), inner sep=0.1cm] (phase2) {};
\node[rotate=90, above=0cm] at (phase2.west) {\textbf{Phase II}};
\node[draw, dotted, line width=0.8pt,fit=(product)(identProdInfo)(extract)(annot)(aufbPM), inner sep=0.1cm] (phase3) {};
\node[rotate=90, above=0cm] at (phase3.west) {\textbf{Phase III}};

\node[draw, dotted,line width=0.8pt, fit=(algorithm)(problem), inner sep=0.1cm] (phase4) {};
\node[ above=0cm] at (phase4.north) {\textbf{Phase IV}};

    \path[line] (sysModel) -- (analyzeSM);
    \path[line] (analyzeSM) -- (scope);
    \path[line] (scope) -- (identRelElements);
   \path[line] (identRelElements) -- ++(0,-1.5cm) -| (genDomain);
   \path[line](profile) -- (genDomain);
   \path[line](genDomain) -- (defAction);
   \path[line](defAction) -- (aufbSM);
   \path[line](aufbSM) -- (genPD);
   \path[line](product) -- (identProdInfo);
   \path[line](identProdInfo) -- (extract);
    \path[line](extract) -- (transmit);
    \path[line](transmit) -- (annot);
    \path[line](annot) -- (aufbPM);
    \path[line](aufbPM) |- ++(0,1.7cm) -| (genPD);
    \path[dashline](genDomain) |- ++(0,-1.5cm)-| (annot);
    \path[line](algorithm)--(genPD);
     \path[line](genPD)  -- (problem);
     \path[line](problem)  -- (solve); 
     \path[line](solve)  -- (plan); 
\end{tikzpicture}
    
    }
    \caption{Workflow Model for Automated Generation of PDDL Descriptions}
    \label{fig:workflow}
\end{figure*}
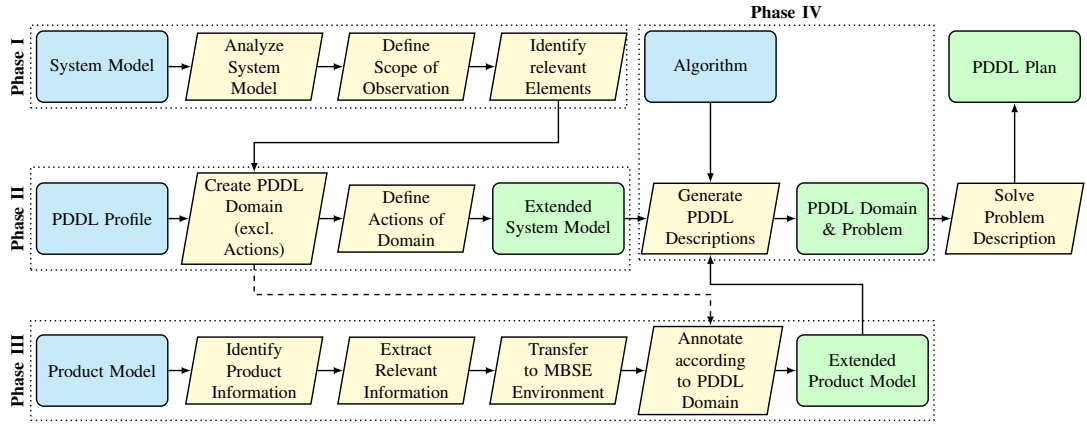

The proposed workflow model facilitates the automated generation of \ac{pddl} descriptions by systematically analyzing and enriching existing system and product models.  
It consists of four phases: \method{Analysis and Preparation of the System Model}, \method{Enrichment of the System Model}, \method{Provision of the Product Model}, and \method{Generation of PDDL Descriptions}.  
The entire workflow is depicted in Fig.~\ref{fig:workflow} and is explained in detail below.

\subsection{Phase I: Analysis and Preparation of the System Model}
The workflow requires a system model as a starting point, typically modeled using the \ac{sysml}. 
A system model offers a comprehensive representation of a system encompassing structural, behavioral, functional, and operational information, as well as constraints, requirements, and quality attributes. 
The use of \ac{sysml} is a prerequisite for the proposed workflow, as it allows for a structured and consistent representation of the system components and their interactions.

Since it cannot be assumed that users of the proposed workflow were involved in developing the system model, the first step is to analyze and understand the existing system model (\method{Analyze System Model}). 
This analysis provides an understanding of the system's architecture and context, ensuring that all relevant components and relations are recognized. 

Afterward, it is essential to define which parts of the system model will be the scope of observation (\method{Define Scope of Observation}). 
For instance, a system model may describe an entire production plant, but only a specific submodule is relevant for the current planning problem. 
This scope must be clearly defined in advance to ensure that subsequent steps focus only on relevant parts of the system. 
By doing so, the planning process avoids unnecessary complexity. 

The next step is to identify which elements within the system model's scope are relevant for the current planning problem~(\method{Identify Relevant Elements}). 
This involves selecting components, interfaces, and relations that play a critical role in the planning problem. 
Identifying these relevant elements establishes a clear mapping between the system model and the planning domain, which helps accurately translate the system's structure into a \ac{pddl} description. 

With a well-defined scope and a clear understanding of the relevant system elements, the next phase focuses on enriching the system model by incorporating annotations and actions required for the \ac{pddl} domain.

\subsection{Phase II: Enrichment of the System Model}
In Phase II, the goal is to enrich the system model with planning-specific aspects. 
This requires the \ac{sysml} Profile for \ac{pddl} \cite{nabizada2024eka} which allows the embedding of \ac{pddl} constructs directly into \ac{sysml} models. 

Therefore, the first step of this phase is the creation of the \ac{pddl} domain by defining the static parts of the domain~(\method{Create PDDL Domain (excl. Actions)}). 
This involves specifying objects, predicates, and types that represent the system's components and their properties. 

Starting with the static parts of the domain is crucial because it establishes the foundational elements and structure of the system. 
Objects and predicates define the components and their relations. 
These static elements are essential because they are reused in the dynamic parts of the domain. 
For example, predicates defined in this step will be used as preconditions and effects in the actions. 
By thoroughly understanding and specifying static elements first, the actions that operate upon these elements can be defined more accurately and effectively. 

Following the definition of static parts, dynamic aspects of the domain are addressed by defining actions that can be performed within the system~(\method{Define Actions of Domain}). 
Actions include preconditions and effects. 
The preconditions outline the conditions under which actions can be executed, and the effects describe the resulting changes in the system's state. 
These preconditions and effects are directly based on objects and predicates defined earlier, ensuring consistency, correctness, and reusability in the planning domain. 

These steps lead to an \textit{Extended System Model} from which targeted information for the PDDL domain description can be queried and extracted. 
 
\subsection{Phase III: Provision of the Product Model}
While the system model provides a wealth of information necessary for creating the \ac{pddl} domain, the product model supplies essential information for defining the \ac{pddl} problem. 
The \ac{pddl} problem describes the specific instance of the planning task, including the initial state and the goal state, which are often closely related to the product's current state and desired outcomes. 
For example, in a manufacturing context, the system model might define the machinery and processes available (\ac{pddl} domain), whereas the product model specifies the current configuration of a product and the desired final configuration (\ac{pddl} problem).

Therefore, in Phase III, the focus shifts to integrating product-specific information into the planning domain. 
The process begins with identifying relevant product information from the existing product model (\method{Identify Product Information}), such as the positions of rivets of an aircraft fuselage. 

After identifying the relevant information, the next step is to extract this information from the product model~(\method{Extract Relevant Information}) and transfer it to the \ac{mbse} environment~(\method{Transfer to MBSE Environment}). 
In the \ac{mbse} environment, the extracted product information is annotated according to the PDDL domain (\method{Annotate according to PDDL Domain}). 
This annotation reuses the previously defined objects, predicates, and types from the \ac{pddl} domain, ensuring that the product model is aligned with the planning domain. 

These steps result in an \textit{Extended Product Model}, enriched with the necessary annotations required for planning.

\subsection{Phase IV: Generation of PDDL Descriptions}

In Phase IV, the focus is on automatically generating a problem description. 
This description can then be submitted to a solver to create a feasible plan based on the integrated system and product models. 
At this point, the generation of \ac{pddl} descriptions, which include both the \ac{pddl} domain and the \ac{pddl} problem file, occurs (\method{Generate PDDL Descriptions}). 
Since the required information has already been prepared in previous phases, an algorithm can directly query this information and translate it into \ac{pddl} syntax. 

Because the system model and the product model are directly associated with planning aspects, generating new \ac{pddl} files can be accomplished with minimal manual effort when changes occur in these models. 
This ensures that the planning problem remains up-to-date and accurately reflects the current state and goals of the system. 

Finally, these generated \ac{pddl} files can be fed into a \ac{pddl} solver, such as those available by the implementation presented in \cite{Muise.2022}, to solve the planning problem (\method{Solve Problem Description}). 
The solver processes \ac{pddl} descriptions to generate a sequence of actions (a plan) that transitions the system from an initial state to a goal state, adhering to all specified constraints and requirements. 

Once the plan is generated, it can be used for further analysis to optimize the system. 
This includes evaluating different scenarios, identifying potential bottlenecks, and exploring alternative strategies to improve efficiency. 
These analyses help in making decisions and ensuring that the system operates at its optimal performance level under specific constraints. 

These steps complete the workflow, resulting in a comprehensive process for generating \ac{pddl} planning problems based on system and product models.

\section{Application Example}
\label{sec:applicationexample}
To validate the developed workflow, a use case from aircraft structure assembly was utilized, featuring a UR10 robotic arm within the fuselage for screwing collars onto rivets, each requiring specific end-effectors. 
The workflow leveraged existing system and product models, developed in \textit{\ac{msosa}} and \textit{3DExperience} respectively, to optimize process flow and reduce throughput times. 
Detailed modeling techniques from \cite{beers2023mbse} and the application of the \ac{pddl} Profile on the system model, as demonstrated in \cite{nabizada2024eka}, facilitated this integration. 
The process involved extracting rivet information from the 3D product model, which was necessary to define the \ac{pddl} problem description. 

For the generation of the PDDL files, \ac{vtl} templates were utilized. 
These templates structure the output of planning information into the standardized \ac{pddl} format. 
By using placeholders dynamically filled by the Velocity Engine integrated within \ac{msosa}, \ac{vtl} templates systematically transform domain-specific data extracted from system and product models into valid \ac{pddl} files. 
This automation ensures that the planning documents align precisely with the specified planning requirements. 

The application of this workflow successfully demonstrated the capability to generate correct \ac{pddl} descriptions from the system and product models and solve them by a \ac{pddl} solver. 
The resulting plan was compared against the expected assembly process to ensure correctness. Actions generated by the solver were checked for logical consistency and alignment with the system and product specifications. 
This validation confirms the workflow's efficacy in automating complex planning tasks, thereby streamlining the production process while ensuring accurate alignment with the objectives of the planning problem. 
This workflow establishes a tight connection between engineering models and the planning problem, allowing for quick adaptation to changes. 
For instance, when a product is modified, new \ac{pddl} files can be generated with minimal effort, ensuring that planning remains aligned with current product specifications. 

\section{Conclusion and Future Work}
\label{sec:conclusionandoutlook}
This contribution outlines a model-based workflow for generating \ac{pddl} descriptions that integrates system and product models. 
This workflow consists of four phases, detailing the steps required to enhance existing system and product models, and automate the generation of \ac{pddl} files. 
Integrating this approach facilitates efficient and flexible planning processes, improving decision-making in complex systems engineering by providing a robust plan that supports engineering decisions and adapts to changes in the models. 
By utilizing \ac{sysml} for system modeling, the approach ensures a structured representation of system components. 
To validate the workflow, a use case from aircraft structure assembly was utilized. 

Unlike the approach by Wally et al. \cite{wally2019flexible}, which is limited to ISA-95 systems and does not integrate with other models, this workflow is designed to be flexible and applicable across various modeling standards. 
This flexibility allows for the integration of different system and product models, making it more adaptable to diverse industrial applications.

However, the workflow faces challenges due to the heterogeneity of models, since modeling standards vary significantly among modelers. 
This is why there are manual efforts in applying this workflow. 
Standardizing the \ac{mbse} modeling workflow could automate these currently manual steps.

While initial applications of the workflow have shown promise in generating correct \ac{pddl} descriptions from system and product models, the improvement in terms of reducing complexity and errors still needs to be thoroughly evaluated. 

Future work will also focus on further refining this workflow and applying it to more complex systems to ensure its applicability in industrial use cases. 
Additionally, the underlying algorithm will be thoroughly documented and published, providing detailed insights into its functionality and implementation. 
This will enable broader application of the workflow across various tools beyond \ac{msosa}, facilitating its use in a wide range of engineering and planning contexts.

\section*{Acknowledgments}
This research paper [project iMOD and LaiLa] is funded by dtec.bw – Digitalization and Technology Research Center of the Bundeswehr. dtec.bw is funded by the European Union – NextGenerationEU.

\bibliographystyle{IEEEtranN}
\bibliography{references.bib}

\end{document}